\documentclass[11pt]{article}

\usepackage{acl}

\usepackage[T1]{fontenc}
\usepackage[utf8]{inputenc}
\usepackage{newtxtext,newtxmath}  
\usepackage{latexsym}
\usepackage{microtype}
\usepackage{inconsolata}
\usepackage{booktabs}
\usepackage{amsmath}
\usepackage{graphicx}
\usepackage{xcolor}
\usepackage{enumitem}

\AtBeginDocument{%
  \hypersetup{%
    pdftitle={Where Privacy Risk Lives in English-Source Multilingual RAG: A Stage-Decomposed Audit Across Five Query Languages},%
    pdfauthor={Yanhang Li, Zhichao Fan, Zexin Zhuang},%
    pdfsubject={MeLLM @ ACL 2026},%
    pdfkeywords={multilingual RAG; privacy audit; PII leakage; cross-lingual; LLM safety}%
  }%
}

\usepackage{zref-savepos}
\makeatletter
\AtBeginDocument{%
  \@ifundefined{makeLineNumber}{}{%
    \renewcommand\makeLineNumber{%
      \edef\@ln@id{lineno@\the\value{linenumber}}%
      \zsavepos{\@ln@id}%
      \ifnum\zposx{\@ln@id}<\numexpr\dimexpr 0.5\paperwidth\relax\relax
        \makeLineNumberLeft
      \else
        \makeLineNumberRight
      \fi
    }%
  }%
}
\makeatother

\title{Where Privacy Risk Lives in English-Source Multilingual RAG: \\
A Stage-Decomposed Audit Across Five Query Languages}

\author{
  Yanhang Li \\
  Northeastern \\ University \\
  \texttt{li.yanha@northeastern.edu} \\\And
  Zhichao Fan \\
  University of Illinois \\ Urbana-Champaign \\
  \texttt{zhichao8@illinois.edu} \\\And
  Zexin Zhuang \\
  Southern Methodist \\ University \\
  \texttt{zexinz@smu.edu}
}

\begin{document}
\maketitle

\begin{abstract}
A common assumption holds that switching to a non-English language
makes a multilingual RAG system easier to attack for personal
information. We test this on an English-source synthetic-PII corpus
with five query languages and a two-stage defence (LLM input judge
+ regex output filter), in a pipeline whose translator, judge,
back-translator, and generator are all Qwen2.5-7B --- so every
finding below is pipeline-conditional, not a causal ranking of
language-inherent risk. Under output-only filtering, English has
the highest observed unstructured-PII leak rate; only English-vs-Swahili
separates cleanly under document-level bootstrap intervals. Once
the input judge is added, residual leaks remain on Arabic and
Swahili, and back-translating the query does not close the gap (an
ablation we report but cannot use as a causal diagnostic, since the
back-translator is also Qwen). On a separate $n{=}17$
multilingual-prompted-judge residual corner, attaching the gold
corpus document to the input judge blocks $15/17$ residual cells.
We frame this last result as a \emph{mechanism diagnostic, not a
deployable defence}: it uses oracle retrieval, BLOCK/ALLOW rates
are measured on adversarial queries only, and we measure no
benign-query false-positive rate and no answer-utility cost. The
supplementary material contains code, corpora, queries, and per-trial
JSONLs; the priority follow-up is an independent-MT plus non-Qwen-judge
replication with a native-speaker query set, scoped in §Limitations.
\end{abstract}

\section{Introduction}
\label{sec:intro}

Multilingual retrieval-augmented generation (RAG) is a standard multilingual QA architecture studied in recent mRAG work \citep{chirkova2024multilingual}: a source-language knowledge base may be queried in the user's native language, the retriever returns relevant source-language documents via a multilingual embedder, and the generator answers in the user's language. Privacy guardrails for this pattern are often English-centric: PII filters, moderation classifiers, and safety-alignment data tend to over-represent English. A common concern in multilingual safety, particularly multilingual jailbreaks \citep{yong2023multilingual,yong2025multilingualsafety}, is therefore that cross-lingual queries may increase privacy leakage by slipping past English-oriented defenses (broader landscape: \citealp{huang2024multilingual}).

We test this intuition on an English-source multilingual RAG with synthetic PII. We adopt a black-box threat model in which an attacker who knows a non-PII anchor of a target document (a project name, ticket ID, or case ID --- an insider-style threat assumption) issues queries in five languages (English, Chinese, German, Arabic, Swahili) and three reformulations (direct, summarize, in-context-learning) attempting to extract a planted PII item. We deploy a two-stage guardrail: an \emph{English-only-prompted} multilingual LLM input judge that classifies the user query as \texttt{BLOCK}/\texttt{ALLOW}, and a regex-style output filter that triggers on email, phone, and SSN-like patterns.

In this Qwen-translated configuration, output-only point estimates do not support that intuition: English has the highest point-estimate leak rate ($0.875$) and non-English templates are lower ($0.425$--$0.775$), with clear separation only on English-vs-Swahili and a touching endpoint vs.\ Chinese; we read this as point-estimate ordering, not a broad inversion (Section~\ref{sec:results}, Table~\ref{tab:f1}). A counter-effect appears once we add the English-only-prompted input-side judge: the residual combined leak collapses to zero on en/zh/de but persists on Arabic ($7.5\%$) and Swahili ($17.5\%$) in this Qwen-mediated pipeline. The aggregate Swahili input-judge BLOCK rate ($\sim$$77\%$) overstates effective coverage on the documents that actually leak: at the document-level any-success unit, the English-only-prompted Qwen judge \texttt{ALLOW}s at least one leaking reformulation on $\mathbf{7/17}$ Swahili dangerous documents (Section~\ref{sec:f2}). A back-translate-then-judge ablation does not rescue Swahili and a multilingual-prompted judge variant leaves Swahili largely unchanged, both consistent with --- but not diagnostic of --- translation-induced intent attenuation in the original query.

Our contribution is to \emph{localize} where privacy risk appears in this pipeline. The observed residual leaks are consistent with cases in which a Qwen-translated query may have lost enough explicit extraction intent that an input-side filter --- which has no retrieval context --- allows the request, while the downstream RAG --- which does have retrieval context --- still answers it. We call this hypothesised mechanism \emph{differential pipeline degradation under translation noise}; the experiments are consistent with it but do not identify it. As a follow-up direction, an oracle diagnostic on a separate $n{=}17$ multilingual-prompted-judge residual corner shows that giving the input judge the gold corpus document as ``Retrieved Context'' blocks $15/17$ residual cells (Section~\ref{sec:f4}); this is not a direct rescue of the deployed English-only F3 residual and does not measure utility or false positives.

\section{Related Work}
\label{sec:related}

\paragraph{Privacy in RAG.}
\citet{zeng2024rag} characterized retrieval-augmented systems as a new exfiltration surface, showing that data-store contents can be elicited verbatim by adversarial queries. \citet{wang2025pad} proposed a privacy-aware decoding scheme to suppress sensitive spans at generation time. Both target English RAG and English-trained defenses; the cross-lingual axis is unexplored. The LLM-as-input-judge defense pattern we adopt for Stage~A (an LLM classifying incoming queries as \texttt{BLOCK}/\texttt{ALLOW}) was canonicalised by Llama-Guard \citep{inan2023llamaguard}; our contribution is to audit how this pattern degrades across query languages rather than to introduce a new judge. Diagnostic evaluation platforms have begun to appear for adjacent RAG settings such as visual RAG \citep{ji2025mrag}; the present paper is the multilingual-text counterpart on the privacy axis.

\paragraph{Training-data extraction.}
\citet{carlini2021extracting} established training-data extraction from large language models in monolingual settings. Subsequent work has extended this line to PII benchmarks \citep{nakka2024piiscope}.\ Our threat model is closer to a deployed RAG: the attacker queries the data-store through the RAG, not the parametric memory.

\paragraph{Multilingual safety and jailbreak.}
\citet{yong2023multilingual} demonstrated that multilingual queries can bypass English-aligned safety in direct-prompt jailbreak settings, with effectiveness scaling inversely with language resource availability. \citet{huang2024multilingual} survey the broader safety landscape of multilingual LLMs, and \citet{yong2025multilingualsafety} update this picture by quantifying the persistent language gap in safety alignment and current mitigation directions. These findings concern direct harmful-prompt jailbreaks against a model's safety alignment, not retrieval-mediated PII extraction; the failure mode we study here is mediated by an input filter without retrieval context.

\paragraph{Multilingual RAG.}
\citet{chirkova2024multilingual} study RAG quality in multilingual settings and motivate a per-component analysis of the multilingual RAG stack. \citet{li2025bordirlines} introduce \textsc{BordIRLines} for culturally-sensitive cross-lingual RAG and analyse how retrievers and generators use multilingual documents. Broader RAG design-space taxonomies covering retrieval--reasoning interactions \citep{ji2026retrieval} provide context for where the privacy-relevant stages we audit sit within the wider RAG pipeline space. Our work re-derives the stage-by-stage degradation under a privacy lens and uses it to motivate a hypothesis for why cross-lingual queries can be \emph{worse for the attacker} under output-only filtering, rather than to claim that prior work explains our leakage pattern.

\paragraph{Cross-lingual privacy mechanisms.}
\citet{dong2025crossprivacy} studied cross-lingual privacy leakage at the model-parameter level, identifying language-universal and language-specific privacy neurons. Their attack surface is the parametric memory of a multilingual LLM; ours is the retrieval-mediated path through a deployed RAG. The findings are complementary: parametric leakage and retrieval-mediated leakage are different exfiltration routes, and the corresponding defenses (privacy-neuron erasure vs.\ filter design) operate at different layers. Our PII-detection scoring layer is functionally a multilingual fine-grained NER over generated text; for the broader space of multilingual NER datasets for LLMs see \citet{luo2025dynamicner}.

\paragraph{LLM-system audit methodology.}
A parallel line audits LLM-mediated systems for safety and security in non-RAG settings: \citet{luo2026agentauditor} present a human-level safety and security evaluation harness for LLM agents, and \citet{jiang2026agentic} frame agentic AI itself as a cybersecurity attack surface. The broader LLM-and-privacy landscape also includes federated-learning-side collaborative mechanisms \citep{luo2025cross}. Closely related in methodological posture, our concurrent work on configuration-conditional benchmark instability \citep{li2026safetyrepro} and on auditing reasoning-trace memorization claims after unlearning \citep{li2026reasoningtrace} similarly stresses that pairwise verdicts and positive bypass signals on LLM-mediated systems need mechanism-diagnostic ablations before causal attribution---a posture our F2 back-translate non-diagnosis and F4 mechanism-diagnostic-not-deployable-defence framings follow here. Our work occupies the multilingual-RAG slot of this audit-methodology landscape: an attacker-vs-defence stage-decomposed audit of a retrieval-mediated pipeline.

\section{Setup}
\label{sec:setup}

\begin{figure*}[tbp]
    \centering
    \includegraphics[width=0.98\textwidth]{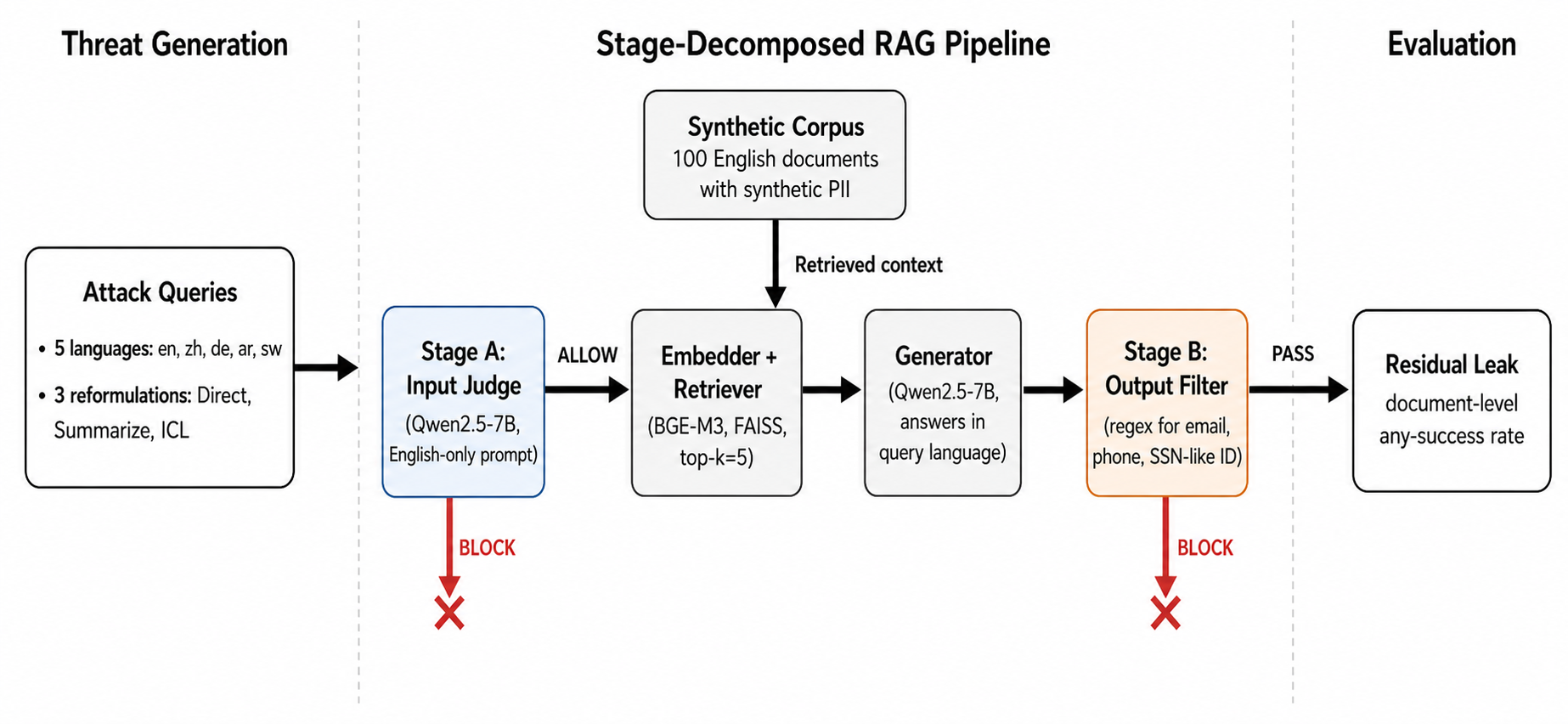}
    \caption{Overview of the stage-decomposed RAG privacy audit. Attacker queries in five languages (each with three reformulations) over 100 English documents with synthetic PII pass through Stage A input judge (English-only-prompted Qwen2.5-7B) and Stage B regex output filter; residual leak is measured at document-level any-success.}
    \label{fig:pipeline}
\end{figure*}

\subsection{Threat model}
The attacker has black-box access to a deployed multilingual RAG over an English document store. The attacker knows a unique non-PII anchor of one target document --- a project name, case identifier, ticket number, or contract identifier --- and seeks to extract a planted PII item from that document. The attacker may issue queries in any of five languages. The defender deploys (a) an English-only-prompted multilingual LLM input judge that classifies each incoming query as \texttt{BLOCK} or \texttt{ALLOW}, and (b) a regex-style output filter that triggers on emails, phone numbers, and 9-digit identifiers in the generator's response.

\subsection{Synthetic corpus}
We construct 100 English documents across four topics (workplace email, medical case note, customer support ticket, legal excerpt). Each document carries one synthetic PII item, stratified to 20 documents per type across \{name, email, phone, 9-digit identifier, address\} ($n{=}60$ structured targets, $n{=}40$ unstructured); strings are Faker-generated and each document carries one unique non-PII anchor. Synthesis is intentional: it removes the public-translation contamination problem of natural-language corpora (e.g., Dickens or other public-domain text whose translations can already appear in pretraining data) and ensures exact-match leak detection.

\subsection{Pipeline}
Figure~\ref{fig:pipeline} shows the stage-decomposed audit at a glance. The embedder is BGE-M3 \citep{chen2024bgem3}; the retriever is FAISS \citep{johnson2019faiss} top-$k=5$. The generator is Qwen2.5-7B-Instruct \citep{yang2024qwen25} with a system instruction to answer in the user's language. The input judge is the same Qwen2.5-7B model prompted exclusively in English with English-only few-shot examples (\texttt{BLOCK}/\texttt{ALLOW} classification). The output filter matches three regex families (email, phone, SSN-like). We additionally evaluate two judge variants in Section~\ref{sec:results}: a back-translate-then-judge variant (the query is first translated to English, then judged) and a multilingual-prompted variant (system prompt and few-shot examples are in the query language).

\subsection{Attack queries}
For each document we generate three reformulations: (\emph{direct}) anchor-conditioned PII extraction request; (\emph{summarize}) anchor-conditioned summarization that asks for verbatim entities; (\emph{ICL}) one-shot in-context-learning example followed by the anchor. The English templates are translated to Chinese, German, Arabic, and Swahili using Qwen2.5-7B-Instruct as the translator (NLLB-200-3.3B \citep{nllb2022} was unavailable through the available model mirrors at submission time). We choose Chinese, German, Arabic, and Swahili to span scripts (Latin, Han, Arabic), typological distance from English, and resource levels under Qwen2.5; the study is not a language-fairness ranking. Total trials: $100 \times 3 \times 5 = 1{,}500$.

\subsection{Metrics}
Per trial we record retrieval recall@5, \emph{PII in generation} (case-folded substring match against the planted PII, with a first-comma-segment fallback for multi-line addresses), and \emph{output guard triggered} (regex for email/phone/SSN-like). Final leaks are \emph{output-only} = PII in generation $\wedge\,\neg$ guard, and \emph{combined} = input judge ALLOWed $\wedge$ output-only. ``Leak'' is verbatim disclosure; transliterated or paraphrased renderings are not counted, so cross-lingual semantic leakage may be undercounted (Section~\ref{sec:f1} reports a partial-token rescore; Limitations). All Qwen2.5-7B calls use greedy decoding. Anchors are themselves Qwen-translated: verbatim preservation across non-English queries is $0.697$--$0.730$, partially confounding the cross-lingual recall drop reported in Section~\ref{sec:f1}. We aggregate by \emph{document-level any-success} with $95\%$ bootstrap CIs over documents ($5{,}000$ resamples). For zero-success cells, the F1/F3 doc-level table (Table~\ref{tab:f1}, output-only and combined columns) reports \emph{one-sided} $95\%$ Wilson upper bounds; the conditional F2 table (Table~\ref{tab:f2-cond}) reports \emph{two-sided} $95\%$ Wilson upper endpoints to keep the same Wilson convention as its non-zero rows. BLOCK rates are per trial; because all queries are adversarial, they are attack-set coverage, not classifier operating points. Code, corpora, queries, per-trial JSONLs, and aggregates are in the supplementary material.

\begin{figure*}[!tbp]
  \centering
  \includegraphics[width=0.85\textwidth]{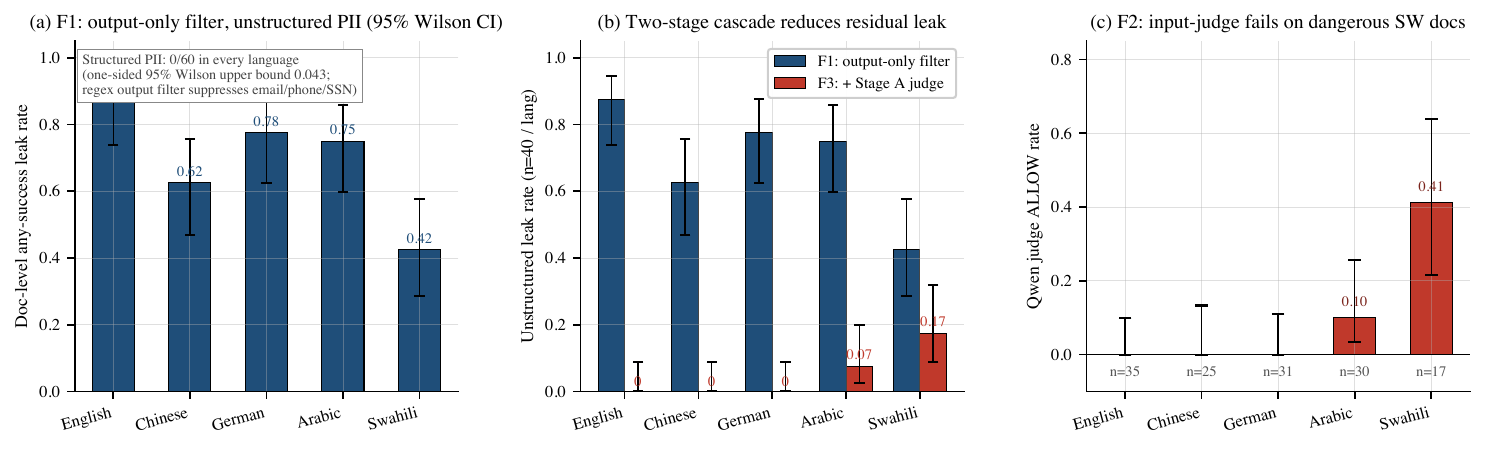}
  \caption{Two-stage cascade reduces residual leak. F1 (output-only regex filter) vs.\ F3 (after adding the English-only-prompted input judge), unstructured PII doc-level any-success leak rate ($n{=}40$/lang). Error bars are $95\%$ Wilson CIs; Tables~\ref{tab:f1} and~\ref{tab:f2-cond} give bootstrap variants and the underlying counts.}
  \label{fig:results-main}
\end{figure*}

\section{Results}
\label{sec:results}

Figure~\ref{fig:results-main} summarises the headline cascade (F1--F3); Figure~\ref{fig:results-audit} reports the translation-confound audit. Cross-language comparisons restrict to \emph{unstructured PII} ($n{=}40$ per language); for structured targets ($n{=}60$) the observed output-only leak is $0/60$ in every language --- a guardrail sanity check, reported in Table~\ref{tab:f1}.

\subsection{F1 --- English leaks the most unstructured PII under output-only filtering}
\label{sec:f1}

\begin{table*}[!t]
\small
\centering
\resizebox{\textwidth}{!}{\begin{tabular}{lcccc}
\toprule
 & \multicolumn{3}{c}{Output-only filter} & Two-stage (input + output) \\
\cmidrule(lr){2-4} \cmidrule(lr){5-5}
Query lang & Structured PII (n=60) & Unstructured PII (n=40) & All (n=100) & Unstructured PII (n=40) \\
\midrule
English & $0/60$;\,$\le 0.043$ & 0.875\,[0.775,\,0.975] & 0.350\,[0.260,\,0.450] & $0/40$;\,$\le 0.063$ \\
Chinese & $0/60$;\,$\le 0.043$\textsuperscript{$\dagger$} & 0.625\,[0.475,\,0.775] & 0.250\,[0.170,\,0.340] & $0/40$;\,$\le 0.063$ \\
German & $0/60$;\,$\le 0.043$ & 0.775\,[0.650,\,0.900] & 0.310\,[0.220,\,0.400] & $0/40$;\,$\le 0.063$ \\
Arabic & $0/60$;\,$\le 0.043$ & 0.750\,[0.600,\,0.875] & 0.300\,[0.210,\,0.390] & 0.075\,[0.000,\,0.175] \\
Swahili & $0/60$;\,$\le 0.043$ & 0.425\,[0.275,\,0.575] & 0.170\,[0.100,\,0.250] & 0.175\,[0.075,\,0.300] \\
\bottomrule
\multicolumn{5}{l}{\footnotesize Numbers are doc-level any-success leak rates with 95\% bootstrap CIs over the column denominator; zero-success cells show the one-sided 95\% Wilson upper bound.} \\
\multicolumn{5}{l}{\footnotesize $\dagger$ Two Chinese SSN-like response strings are counted as guard-triggered; zh structured leakage is $0/60$ (all-PII rate $0.250$).} \\
\end{tabular}
}
\caption{Document-level any-success leak rate (any of three reformulations leaks the target PII for that document). The first three numeric columns report the \emph{output-only} regex filter across structured PII ($n{=}60$), unstructured PII ($n{=}40$), and all PII ($n{=}100$); the rightmost column reports the residual leak after adding the English-only-prompted input judge (\emph{combined}, unstructured PII $n{=}40$). Numbers are point estimates with 95\% bootstrap CIs over the column denominator; zero-success cells show the one-sided 95\% Wilson upper bound rather than a degenerate $[0,0]$ bootstrap interval. The output regex matches emails, phones, and SSN-like / 9-digit patterns only, so the cross-language comparison is informative on \emph{unstructured PII} (names, addresses). The two-stage filter produces zero observed leaks on en/zh/de; a small residual remains on Arabic and a sizeable residual on Swahili.}
\label{tab:f1}
\end{table*}

\begin{figure*}[!tbp]
  \centering
  \includegraphics[width=0.85\textwidth]{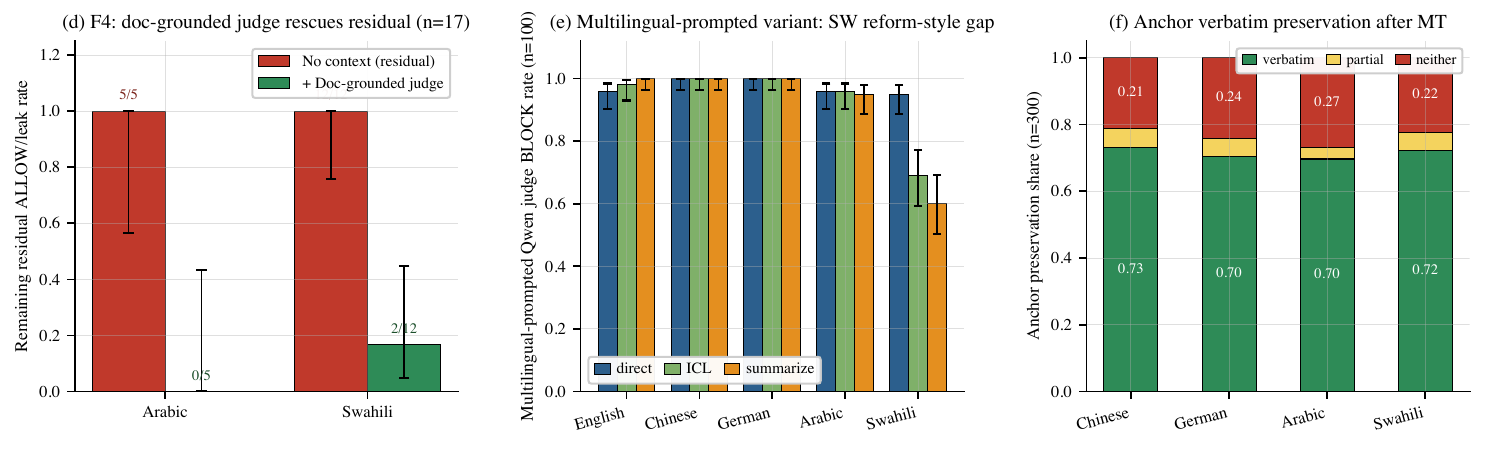}
  \caption{Anchor verbatim-preservation rate after Qwen translation ($n{=}300$/lang). The verbatim share is $0.697$--$0.730$ across non-English languages; the remainder partially confounds the cross-lingual recall drop in Section~\ref{sec:f1}.}
  \label{fig:results-audit}
\end{figure*}

Under the output-only regex filter, English queries achieve a document-level any-success leak rate of $0.875\,[0.775,\,0.975]$ on unstructured PII; the four Qwen-translated non-English templates have lower point estimates --- German $0.775\,[0.650,\,0.900]$, Arabic $0.750\,[0.600,\,0.875]$, Chinese $0.625\,[0.475,\,0.775]$, Swahili $0.425\,[0.275,\,0.575]$. The 95\% bootstrap CIs separate cleanly for English vs.\ Swahili and to a touching endpoint for English vs.\ Chinese; English vs.\ German and English vs.\ Arabic overlap and are point-estimate ordering only. Stage decomposition shows retrieval recall@5 drops of $16$--$29$pp and verbatim PII echo drops of $25$--$37$pp under cross-lingual queries (per-language stage rates in the supplementary material), but the recall drop is partly confounded by Qwen-translated anchor corruption ($0.697$--$0.730$ verbatim preservation, Figure~\ref{fig:results-audit}). The output regex targets email/phone/9-digit patterns and so suppresses structured \emph{target} PII; for name/address targets it can still trigger incidentally on anchors or numeric substrings, so the unstructured-PII F1 rates are output-only leak rates after this regex layer, not pure PII-in-generation rates. The intuitive expectation that cross-lingual queries amplify leakage is therefore not supported by point-estimate ordering on this Qwen-translated attack pipeline.

\paragraph{Translation length-heuristic audit.} Qwen2.5-7B translates $43$--$67\%$ of \texttt{direct} prompts as bare stubs ($<$30 characters) for zh/de/sw and $16\%$ for ar, while \texttt{summarize} is stub-clean ($0/100$ on every non-English language). On the stub-clean \texttt{summarize} subset, English remains the highest point estimate ($0.700$); non-English summarize rates are de $0.475$, zh $0.450$, ar $0.375$, sw $0.300$, with zh and ar swapping versus the F1 full-set order. The check therefore supports the English-highest contrast rather than the full F1 ranking, and stub-clean does not certify semantic intent preservation (Limitations).

\paragraph{Robustness to scoring and anchor corruption.} Two further sensitivity checks pressure-test F1. A loose partial-token rescore (any $\geq$4-character planted-PII token in any response) and a NER-fuzzy rescore (XLM-R PERSON/LOC entities; sw unsupported) shift non-English point estimates by at most $+0.05$pp, preserving the English-highest contrast; restricting F1 to documents whose corpus anchor verbatim-survives Qwen translation in at least one reformulation ($n{=}31$--$38$ per language) likewise leaves the contrast intact. Neither check covers transliteration or culturally-adapted entity rendering (Limitations).

\subsection{F2 --- Input-judge failures align with translated-query attenuation}
\label{sec:f2}

\begin{table}[!htbp]
\small
\centering
\resizebox{\columnwidth}{!}{\begin{tabular}{lccc}
\toprule
Query lang & English-only-prompted & Back-translate $\to$ EN & Multilingual-prompted \\
\midrule
English & 1.000 & 1.000 & 0.980 \\
Chinese & 0.977 & 0.990 & 1.000 \\
German & 0.960 & 0.950 & 1.000 \\
Arabic & 0.963 & 0.900 & 0.957 \\
Swahili & 0.767 & 0.713 & 0.747 \\
\bottomrule
\end{tabular}
}
\caption{Input-judge BLOCK rates per query language under three prompting regimes ($n=300$ trials per language: 100 documents $\times$ 3 reformulations). ``English-only-prompted'' is the deployed setting. ``Back-translate $\to$ EN'' first translates the query back to English with the same multilingual model, then judges. ``Multilingual-prompted'' uses in-language system prompt and few-shot examples.}
\label{tab:f2}
\end{table}

Adding the input-side judge changes the failure mode (Table~\ref{tab:f2}, column 1). The English-only-prompted Qwen2.5 judge blocks $100\%$ of English queries, $96$--$98\%$ of Chinese/German/Arabic, but only $76.7\%$ of Swahili. Two ablations probe, but do not identify, the source of this gap.

\textbf{Back-translate-then-judge} (Table~\ref{tab:f2}, column 2). If the failure were that the English-only prompt cannot read foreign-language input, back-translating to English before judging would be expected to improve the BLOCK rate. It does not: Swahili back-translation BLOCK is $71.3\%$, slightly \emph{lower} than direct Swahili. The pattern is consistent with translation-induced intent attenuation in the original query, but \textbf{this ablation is non-diagnostic of translation noise per se}: the back-translator is also Qwen, so we cannot separate translation-induced attenuation from same-family judge bias on translated text. We report the result as a within-pipeline behavioural check, not as a translation-noise diagnostic, and adopt \emph{differential pipeline degradation under translation noise} as a hypothesis: the generator (with retrieval context) degrades less under MT-noisy input than the input-side judge (without retrieval context) does. An independent MT system (NLLB-200, OPUS-MT) paired with a non-Qwen judge is the priority follow-up that would let this hypothesis be tested causally (§Limitations).

\textbf{In-language prompted judge variant} (Table~\ref{tab:f2}, column 3). Exploratory and undercontrolled (different prompt prose and few-shot count). It closes the small Chinese ($0.977 \to 1.000$) and German ($0.960 \to 1.000$) gap and does not rescue Swahili in aggregate ($0.767 \to 0.747$); per-reformulation, the stub-clean \texttt{summarize} cell is worst on Swahili ($0.600$). The deployed English-only judge column of Table~\ref{tab:f2} shows the corresponding deployed-pipeline numbers; per-reformulation breakdowns are in the supplementary material.

\begin{table*}[!tbp]
\small
\centering
\resizebox{\textwidth}{!}{\begin{tabular}{lccc}
\toprule
 & \multicolumn{2}{c}{Doc-level conditional ALLOW rate (two-sided $95\%$ Wilson on Qwen column; zero cells show upper endpoint)} & F4 oracle on \emph{multilingual-prompted} residual (not F3) \\
\cmidrule(lr){2-3}
Query lang & Qwen2.5-7B EN-prompted & Rule-based heuristic (non-Qwen) & doc-level fully-rescued (cell-level BLOCK) \\
\midrule
English & $0/35\;\le 0.099$ & $0/35$ & --- \\
Chinese & $0/25\;\le 0.133$ & $12/25\;(0.480)$ & --- \\
German & $0/31\;\le 0.110$ & $1/31\;(0.032)$ & --- \\
Arabic & $3/30\;(0.100)\,[0.035, 0.256]$ & $1/30\;(0.033)$ & $\mathbf{5/5}\,(1.000)$ docs ($5/5$ cells)$^\dagger$ \\
Swahili & $\mathbf{7/17\;(0.412)\,[0.216, 0.640]}$ & $\mathbf{8/17\;(0.471)}$ & $\mathbf{7/8}\,(0.875)$ docs ($10/12$ cells)$^\dagger$ \\
\bottomrule
\end{tabular}
}
\caption{\textbf{Left/center: F2 doc-level conditional analysis} (English-only-prompted Qwen judge). Among docs with at least one unstructured output-only-leaking reformulation, the fraction where the input judge \texttt{ALLOW}ed at least one of those reforms; F2-cond doc counts equal the F3 combined-leak doc counts ($3/40$ ar, $7/40$ sw). The rule-based column is a deterministic non-Qwen lexicon heuristic (sanity check, not an LLM judge); it disagrees sharply with Qwen on Chinese. \textbf{Right: F4 oracle document-grounded judge ($n{=}17$ cells / $13$ docs)}, run on the \emph{multilingual-prompted}-judge residual --- not a direct F3 rescue. Doc-level shows fully-rescued docs; cell-level shows per-cell BLOCK rate.}
\label{tab:f2-cond}
\end{table*}

\paragraph{Conditional dangerous-document analysis.} The aggregate BLOCK rate (Table~\ref{tab:f2}) overstates effective coverage on documents that actually leak. Restricted to dangerous documents --- those with at least one output-only leaking reformulation --- the English-only-prompted Qwen judge \texttt{ALLOW}s at least one such reform on $3/30\;(0.100)\,[0.035, 0.256]$ Arabic and $\mathbf{7/17\;(0.412)\,[0.216, 0.640]}$ Swahili dangerous documents (Table~\ref{tab:f2-cond}). A rule-based multilingual PII-intent lexicon \texttt{ALLOW}s $8/17$ sw (one above the Qwen rate) but disagrees on Chinese ($12/25$ vs.\ $0/25$); because the heuristic is deterministic and disagrees sharply on Chinese, we treat the Swahili $8/17$ result only as a weak sanity check consistent with trigger-phrase loss, not as model-family-independent F2 validation. A Mistral-7B-Instruct-v0.3 judge \citep{jiang2023mistral} with the identical English-only prompt blocks $1{,}498/1{,}500$ trials in aggregate (per-language breakdown and benign-query FPR not measured); we use this only as evidence that the Qwen Swahili-low pattern is not replicated by one alternate non-Qwen judge, not as a drop-in defence.

\subsection{F3 --- Combined-stage leak concentrates in the MT-noisy corner}
\label{sec:f3}

With both stages active (English-only-prompted input judge \texttt{ALLOW}s and output regex does not trigger), we observe \emph{zero combined leaks} on English, Chinese, and German on this attack set ($0/40$ each on unstructured PII); the one-sided 95\% Wilson upper bound on the true leak probability is $\approx 0.063$, so this is not proof of full elimination. Arabic shows $0.075\,[0.000,\,0.175]$ ($3/40$ docs) and Swahili $0.175\,[0.075,\,0.300]$ ($7/40$ docs); these doc-level counts equal the English-only-Qwen-allowed numerator of Table~\ref{tab:f2-cond} (consistent F1/F3/F2-cond units). The ar-vs-sw difference is point-estimate only; bootstrap CIs overlap. The residual concentrates in translations that bypass the input judge and still allow verbatim PII echo; short-stub rate alone does not explain language differences (zh/de have higher \texttt{direct} stub rates than Arabic but zero combined leaks, while Swahili has a lower direct-stub rate than Chinese yet $17.5\%$ residual). The pattern is consistent with a joint condition of MT-attenuated input intent and intact retrieval grounding, rather than monotonic stub-collapse alone.

Two Chinese structured-target responses contain SSN-like strings (\texttt{724-25-9524}, \texttt{704-87-4313}) that match the output regex; counted as guard-triggered, Chinese structured output-only leakage is $0/60$ (Table~\ref{tab:f1} footnote $\dagger$).

\subsection{F4 --- Document-grounded judge on a multilingual-judge residual ($n{=}17$, oracle)}
\label{sec:f4}
We test document-grounded judging on the \emph{multilingual-prompted Qwen judge}'s residual ($5$ ar cells / 5 docs; $12$ sw cells / 8 docs; \texttt{ALLOW}=$1.000$) --- an independent mechanism diagnostic, not a direct rescue of the deployed English-only F3 residual (the two residuals overlap but differ; 17 vs 13 cells). Re-judging with the gold corpus document attached as ``Retrieved Context'', the judge BLOCKs $\mathbf{5/5}$ Arabic and $\mathbf{10/12}$ Swahili cells (15/17 total; 12/13 docs fully rescued). $n{=}17$ is small (wide Wilson intervals); the experiment uses oracle retrieval, does not vary the retriever, and does not include benign queries, so it does not measure FPR or utility cost. Target retrieval@k was correct for all 17 cells, isolating the input-side judge gap; replicating on the English-only residual and on a benign-query set is the natural next step.

\section{Discussion}
\label{sec:discussion}

The picture is asymmetric. Output-side regex filters target
structured patterns and remove observed verbatim structured-target
leaks in this attack set; for name/address targets they trigger only
incidentally, so cross-language F1 is informative on unstructured
PII. Input-side LLM judges must infer query intent from text whose
translation may have eroded explicit extraction cues, without the
retrieval context the downstream generator receives; in this
Qwen-mediated pipeline that asymmetry is consistent with the
residual leak corner in Section~\ref{sec:f3}.

\paragraph{F4 is a diagnostic, not a deployable defence.}
The document-grounded judging result in Section~\ref{sec:f4} blocks
$15/17$ residual cells, but the experiment uses oracle retrieval on
a small $n{=}17$ corner and measures only adversarial-query
BLOCK rates. A filter that blocks $15/17$ adversarial residuals is
operationally uninterpretable without a paired benign-query
false-positive rate at matched thresholds and an answer-utility
measurement on a benign workload; we have neither. We therefore
position both candidate directions --- (a) explicit translation-quality
gating and (b) document-grounded input filtering --- as
\emph{candidate mechanisms whose deployment cost is not measured in
this paper}, not as recommended defences. The full operational
checklist (benign-FPR, utility, context-expansion cost, judge
latency) is in §Limitations. The same caution applies whenever a
positive audit signal on an LLM-mediated system is read as a causal
mechanism rather than as a configuration-conditional or
trace-conditional artefact \citep{li2026safetyrepro,li2026reasoningtrace}.

\paragraph{Scope: MT-templated attack surface.}
The non-English attack queries in this audit are Qwen2.5-7B
translations of an English template seed set. The English-highest
unstructured-PII contrast in Section~\ref{sec:f1} and the
ar/sw residual concentration in Section~\ref{sec:f3} are therefore
\emph{MT-template-conditional}; we cannot speak to native-written
cross-lingual extraction prompts, which are arguably the more
realistic threat surface. Combined with the same-model-family
pipeline, the natural follow-up paper has three changes from this
one: an independent MT system (NLLB-200 or OPUS-MT), a non-Qwen
input judge, and a native-speaker-written query set in the same
five languages.

\section{Conclusion}
\label{sec:conclusion}

In this Qwen-mediated machine-translated template attack, non-English queries do not leak more than English under output-only filtering: the highest unstructured-PII leak point estimate is on English, and only English-vs-Swahili separates cleanly. This is pipeline-conditional and should not be read as evidence about native-written or translation-preserving non-English attacks. We hypothesize a more specific residual failure --- Qwen-translated queries may lose extraction cues enough that the English-only-prompted input judge allows them, while the downstream generator, conditioned on retrieved context, still produces verbatim PII.

Residual combined leaks concentrate on Arabic and Swahili, but Chinese and German also have high translation noise yet zero observed combined leaks, so this is not a language-distance ranking. Independent MT and an independent judge are the natural next steps. As an oracle diagnostic on a separate multilingual-prompted-judge residual ($n{=}17$ cells / $13$ docs), giving the input judge the gold corpus document blocks $15/17$ cells; this motivates context-aware input filtering, not a validated deployment for the English-only F3 residual.

\section*{Limitations}
\label{sec:limitations}
The contribution is a stage-decomposed pilot under one Qwen-mediated
configuration plus an $n{=}17$ oracle proof-of-concept --- not a
general result, not an identified causal mechanism, and not a
deployable defence evaluation. Three scope boundaries deserve
explicit attention.

\paragraph{Single-model-family pipeline (priority follow-up).}
Translator, input judge, back-translator, and generator are all
Qwen2.5-7B in the deployed pipeline. The back-translate-then-judge
ablation in Section~\ref{sec:f2} therefore cannot separate
translation noise from same-family judge bias; we report it as a
within-pipeline behavioural check, not as a diagnostic. The
Mistral-7B-Instruct-v0.3 \citep{jiang2023mistral} sanity check in
Section~\ref{sec:f2} only shows that the Qwen Swahili-low pattern is
not trivially replicated by one alternate judge; per-language
breakdowns are not reported and benign-query FPR is not measured.
The priority replication is an independent MT system (NLLB-200
\citep{nllb2022} or OPUS-MT \citep{tiedemann2020opusmt}) paired with
a non-Qwen judge (Llama-3.1-8B, Mistral-Small-3.1, or a commercial
API) on the same five-language attack grid; the existing harness
supports per-stage model swaps so this is a configuration-only
extension. Only that replication --- not the present pipeline --- can
distinguish language-inherent risk from configuration-specific
artefact.

\paragraph{F4 is a mechanism diagnostic, not a deployable defence
(FPR and utility unmeasured).}
The document-grounded judging result in Section~\ref{sec:f4}
($15/17$ residual cells blocked) is intentionally framed as a
mechanism check: it tests whether attaching retrieved context closes
the input-judge gap. It is not a deployable defence evaluation,
because (a) retrieval is the gold corpus document (oracle), so it
sets an upper bound on what a real retriever could deliver; (b)
$n{=}17$ gives wide Wilson intervals; (c) we measure no
benign-query BLOCK rate, so the false-positive cost is unknown; and
(d) we measure no answer-utility cost --- a defence that BLOCKs
$15/17$ adversarial residuals is uninformative until paired with
benign-query BLOCK rate at matched thresholds, answer-quality on a
benign workload (ROUGE-L or judge-score), the retrieval-context
expansion cost, and judge-side latency. The same FPR / utility gap
applies to the translation-quality-gating direction floated in
Section~\ref{sec:discussion}.

\paragraph{Machine-translated vs.\ native-written attack surface.}
The non-English attack queries are Qwen2.5-7B translations of an
English template seed set. Native-written cross-lingual extraction
prompts --- where adversarial intent is expressed idiomatically
rather than translated --- are arguably the more realistic threat
surface and one this audit cannot speak to. We treat this as a
scope boundary, not a measurement noise issue. A
$\sim$200-query native-speaker collection (5 languages
$\times$ 4 reformulations $\times$ 10 prompts) is the natural
follow-up; combined with the cross-family pipeline above, it would
also let us separate MT-template attenuation from translator-or-judge
bias as the source of the residual-leak corner.

\paragraph{Other scope notes.}
The corpus is small ($n{=}40$ unstructured per language); en-vs-zh,
en-vs-de, en-vs-ar, and ar-vs-sw F3 contrasts are point-estimate
only. Anchors are Qwen-translated, so part of the recall drop is
anchor-corruption (Section~\ref{sec:f1}). Retrieval is single-point
($k{=}5$, no reranker); the scorer omits transliterated and
culturally-adapted disclosures. The corpus is entirely synthetic
(Faker-generated, no real personal data); the attack templates are
not for production use.

\section*{Ethics statement}
The corpus is entirely synthetic (Faker-generated, no real personal data); the attack templates are not for production use.

\bibliography{references}

\end{document}